\documentclass[10pt,twocolumn,letterpaper]{article}

\usepackage{cvpr}              
\definecolor{cvprblue}{rgb}{0.21,0.49,0.74}
\usepackage[pagebackref,breaklinks,colorlinks,allcolors=cvprblue]{hyperref}

\usepackage{microtype,multirow,colortbl}
\DeclareMathOperator*{\argmax}{arg\,max}
\newcommand{\tiou}{\operatorname{tIoU}}
\newcommand{\clip}{\operatorname{clip}}

\def\confName{CVPR}
\def\confYear{2027}

\title{Learning High-Quality Pseudo-Labels for Point-Supervised Temporal Action Localization}

\author{Zhenying Fang\\
{\tt\small zhenying.fang@foxmail.com}
}

\begin{document}
\maketitle
\begin{abstract}
Point-supervised temporal action localization (PTAL) aims to recognize and localize action instances in untrimmed videos using only point-level annotations. Existing pseudo-label-based methods have made substantial progress in PTAL. However, confidence-based initialization overlooks well-localized proposals. The resulting pseudo-labels are then either kept fixed during detector training or refined without fully exploiting the original proposal pool, limiting supervision quality and detection performance. To learn high-quality pseudo-labels, we propose a framework comprising Prior-Guided Initialization (PGI) and Consensus-Guided Refinement (CGR). PGI combines proposal confidence with point-derived priors to establish instance-specific reliability for initial pseudo-label selection and candidate retention. CGR continually uses the retained original candidates to construct boundary targets, combining their consensus with evolving detector predictions to refine pseudo-labels throughout training. Experiments on four benchmarks demonstrate that our method achieves competitive detection performance compared with existing methods. In particular, our method achieves 68.5\% average mAP over tIoU thresholds 0.1--0.7 on THUMOS14, surpassing PseR by 3.7 percentage points.
\end{abstract}    
\section{Introduction}
\label{sec:intro}

\begin{figure}[t]
\centering
\includegraphics[width=\linewidth]{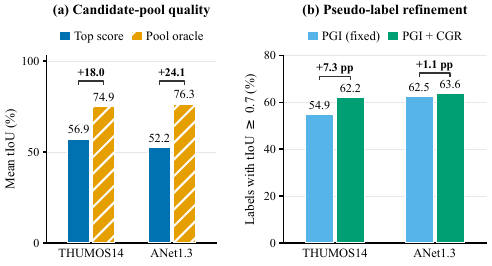}
\caption{\textbf{Proposal quality and pseudo-label refinement.} (a) Mean tIoU of top-score and oracle (maximum ground-truth tIoU) proposals. (b) Fraction of pseudo-labels with tIoU $\geq 0.7$ for fixed PGI and PGI+CGR. Brackets show gains in percentage points.}
\label{fig:gap}
\end{figure}

Temporal action localization (TAL) aims to recognize action instances and determine their temporal boundaries in untrimmed videos~\cite{zhang2022actionformer,fang2023bdrcnet}, supporting applications such as visual question answering~\cite{lei2018tvqa}, video summarization~\cite{lee2012discovering}, and highlight detection~\cite{xiong2019less}. Fully-supervised methods~\cite{shi2023tridet,fang2019pcpcad,wang2026mdvlmtal} have achieved impressive performance, but require precise temporal boundaries for individual action instances, which are costly and labor-intensive to annotate. This annotation burden has motivated weakly-supervised temporal action localization (WTAL)~\cite{ren2023pmil,rizve2023pivotal}, which learns from video-level category labels. Without instance-level temporal cues, however, recovering complete action intervals remains difficult, leaving a substantial performance gap to fully-supervised methods.

To balance detection performance and annotation cost, point-supervised temporal action localization (PTAL) has been proposed as an alternative supervision setting. PTAL requires only one annotated frame and its category per action instance. Early methods~\cite{ma2020sfnet,lee2021lacp,yin2023proposal} follow a \emph{localization-by-classification} paradigm, inferring action intervals from Class Activation Sequences (CAS) learned under point supervision. However, CAS are learned using classification objectives, and the mismatch between classification and localization limits the detection performance of these methods~\cite{fu2022crrcnet,fang2026pser}. To provide explicit localization supervision, \emph{pseudo-label-based} methods construct temporal pseudo-labels to train a detector~\cite{zhang2023hrpro,fang2023lpr,fang2026pser}. For example, POTLoc~\cite{vahdani2024potloc} generates pseudo-labels using point-compatible proposals and proposal statistics. PseR~\cite{fang2026pser} propagates seed proposals through temporal perturbations and refines pseudo-labels using proposal selection and ranking.

Despite this progress, pseudo-label quality remains constrained at both initialization and subsequent detector training. First, proposal confidence does not necessarily reflect localization quality for the annotated instance. Fig.~\ref{fig:gap}(a) reveals a substantial gap between top-score and oracle selection from the same point-compatible candidate pool, showing that well-localized proposals can be overlooked during initialization. Second, pseudo-labels either remain fixed during detector training or are refined without fully exploiting alternative boundaries from the original pool. Fixed labels cannot incorporate new localization evidence, and retaining only the selected pseudo-label intervals excludes other candidate boundaries from subsequent supervision. Together, these limitations restrict supervision quality and the detector's ability to learn accurate action boundaries.

Our central idea is to use a shared assessment of candidate reliability to select initial pseudo-labels and aggregate alternative proposal boundaries during detector training. We implement this idea through two modules: Prior-Guided Initialization (PGI) and Consensus-Guided Refinement (CGR). PGI assesses the reliability of candidate proposals associated with each annotated point by combining confidence, geometric, and conflict priors. Using these reliability scores, PGI selects initial pseudo-labels while retaining alternative candidates with their original boundaries and reliability scores to support subsequent pseudo-label refinement.

CGR reuses the candidates retained by PGI to progressively refine pseudo-label boundaries during detector training. At each update round, it combines a reliability-weighted consensus of these candidates with current detector predictions to construct boundary targets, which guide consistency-controlled pseudo-label updates. The resulting refinement increases the fraction of pseudo-labels with tIoU $\geq 0.7$ by 7.3 and 1.1 percentage points over fixed PGI supervision on THUMOS14 and ActivityNet1.3, respectively (Fig.~\ref{fig:gap}(b)).

Our main contributions are summarized as follows:
\begin{itemize}
    \item We introduce PGI to assess candidate reliability using point annotations, enabling instance-aware pseudo-label initialization while retaining alternative boundary hypotheses for subsequent refinement.
    \item We introduce CGR to construct boundary targets by combining consensus from retained original candidates with current detector predictions, enabling alternative boundary evidence to guide iterative pseudo-label refinement.
    \item Experiments on THUMOS14, ActivityNet1.3, GTEA, and BEOID demonstrate improved pseudo-label quality and competitive detection performance.
\end{itemize}
\section{Related Work}
\label{sec:related_work}

\subsection{Temporal Action Localization}

Fully-supervised temporal action localization (TAL) relies on instance-level category labels and precise temporal boundary annotations. Representative detectors follow anchor-based, anchor-free, hybrid, or query-based designs. Anchor-based methods~\cite{lin2017single,xu2017rc3d,yang2020a2net} use predefined temporal windows, either predicting categories and boundary offsets in one stage~\cite{lin2017single} or generating proposals for subsequent refinement and classification in two stages~\cite{xu2017rc3d}. Temporal anchors provide useful duration priors, but their predefined scales can restrict flexibility when action durations vary substantially~\cite{yang2020a2net}. To alleviate these restrictions, anchor-free methods~\cite{zhang2022actionformer,shi2023tridet,agrawal2025adatadpp,gao2026lsgnet} eliminate manually designed temporal anchors. They predict action categories and distances to temporal boundaries directly at temporal locations. Hybrid approaches~\cite{fang2019pcpcad,yang2020a2net} further integrate anchor-based and anchor-free branches to exploit their complementary strengths. Across these dense prediction designs, post-processing such as Non-Maximum Suppression (NMS) is typically used to remove duplicate detections. Query-based detectors~\cite{liu2022tadtr,kim2025digit} instead formulate TAL as set prediction with learnable action queries, using one-to-one matching during training to discourage duplicate detections. Despite these architectural advances, training still relies on costly temporal boundary annotations, motivating localization under weaker supervision.

\subsection{Weakly-Supervised Temporal Action Localization}

Weakly-supervised temporal action localization (WTAL) reduces annotation costs by learning from video-level category labels without temporal boundary annotations~\cite{yang2022acgnet,li2023boosting,ren2023pmil}. A common approach follows the \emph{localization-by-classification} paradigm~\cite{yang2022acgnet,yang2025crc,shao2026modeling}, which trains a classifier to estimate snippet-level class scores and derives action intervals by thresholding the resulting activation sequences. However, classification objectives tend to emphasize the most discriminative snippets, while less distinctive parts of an action may receive weak responses. This discrepancy makes it difficult to recover complete action instances from classification scores alone.

To address the mismatch between classification and localization, \emph{pseudo-label-based} methods~\cite{rizve2023pivotal,yun2023weakly,liu2025pseudoformer,li2026aclnet} generate temporal pseudo-labels to train a detector. However, constructing accurate pseudo-labels remains challenging because video-level labels do not specify instance positions. Point annotations provide sparse instance-level temporal cues to guide pseudo-label construction.

\begin{figure*}[t]
\centering
\includegraphics[width=\textwidth]{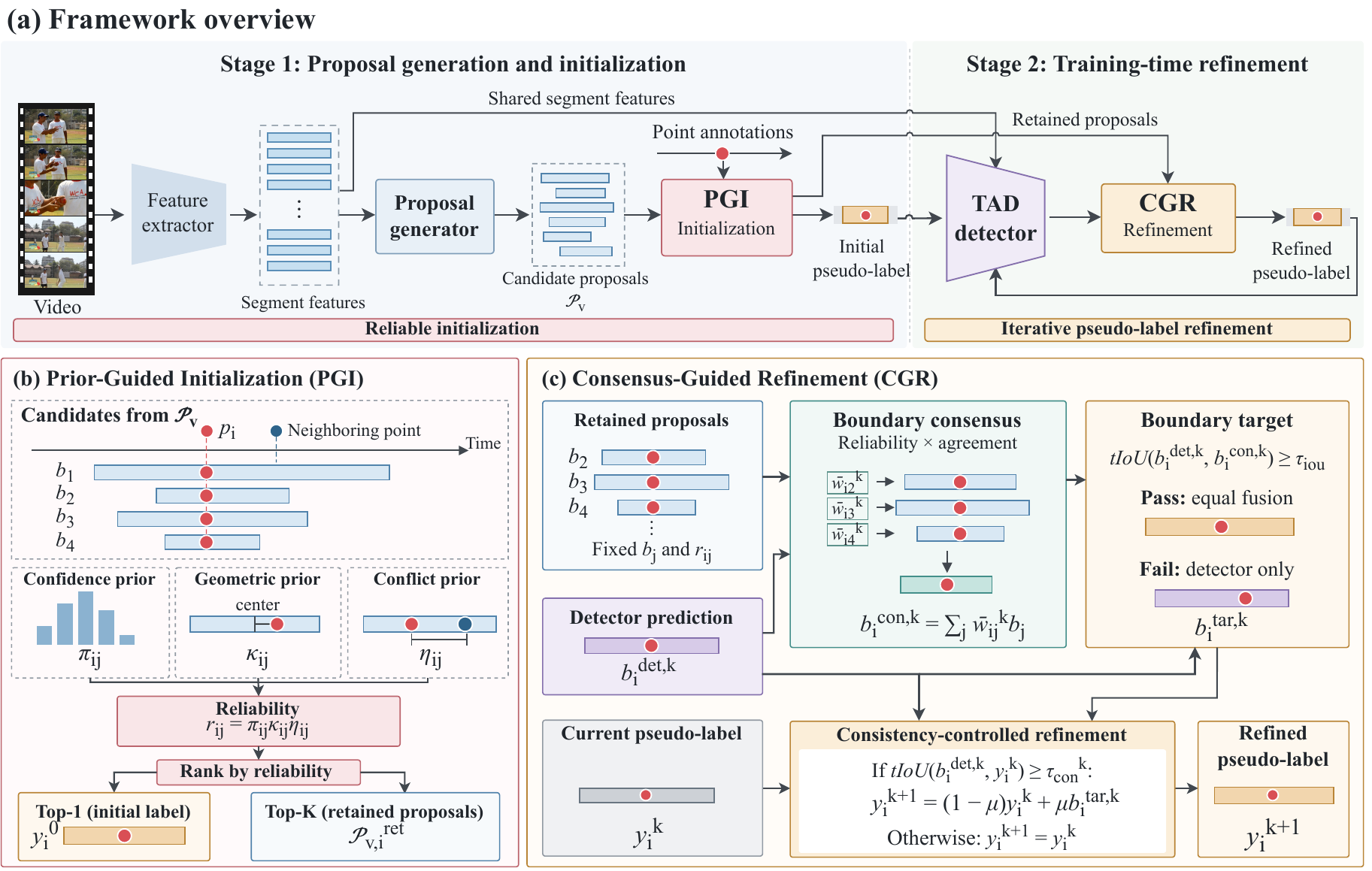}
\caption{\textbf{Overview of our pseudo-label learning framework.} (a) Proposal generation and initialization, followed by detector training with iterative pseudo-label refinement. (b) PGI combines confidence, geometric, and conflict priors into candidate reliability scores for initial pseudo-label selection and top-$K$ retention. (c) CGR combines consensus from retained original proposals with current detector predictions to construct boundary targets, then refines pseudo-labels through consistency-controlled momentum updates.}
\label{fig:framework}
\end{figure*}

\subsection{Point-Supervised Temporal Action Localization}

Point-supervised temporal action localization (PTAL) requires only a single annotated frame and its category for each action instance~\cite{ma2020sfnet}. Early \emph{localization-by-classification} methods expand sparse point supervision and strengthen action representations. SF-Net~\cite{ma2020sfnet} mines labels for neighboring snippets, while LACP~\cite{lee2021lacp} learns action completeness through pseudo-label sequence search. CRRC-Net~\cite{fu2022crrcnet} improves representation compactness and snippet-label reliability, while proposal-level learning~\cite{yin2023proposal} generates dense supervision through clustering. Related efforts model temporal structure through auxiliary tasks~\cite{pointma2026exploring} or enrich visual representations with refined textual descriptions~\cite{ma2027tra}.

A complementary line of research generates temporal pseudo-labels from point annotations to train a detector~\cite{fang2023lpr,zhang2023hrpro,fang2026pser,vahdani2024potloc,zhang2025noco,liu2025qrot}. POTLoc~\cite{vahdani2024potloc} refines point-supervised proposals into pseudo-labels for self-training. NoCo~\cite{zhang2025noco} introduces online label revision through a teacher--student framework and is also evaluated under point supervision.

We focus on retaining and reusing original proposal boundaries across pseudo-label initialization and detector training. Shared candidate reliability guides both initial pseudo-label selection and the retention of alternative proposals. Their original boundaries remain fixed and contribute to a reliability-weighted consensus, which is combined with current detector predictions to construct boundary targets for iterative pseudo-label refinement.
\section{Method}
\label{sec:method}

\subsection{Problem Formulation and Overview}
\label{subsec:3.1}

Given an untrimmed video $v$ of duration $T_v$, point supervision provides annotations $\mathcal A_v=\{(p_i,c_i)\}_{i=1}^{N_v}$, where $N_v$ is the number of annotated action instances, $p_i$ is a timestamp within the $i$-th instance, and $c_i$ is its category. Instance boundaries are unavailable during training. Our goal is to construct temporal pseudo-labels from these annotations and train a detector to recognize and localize action instances in unseen videos.

Following the setup of existing \emph{pseudo-label-based} methods~\cite{fang2023lpr,fang2026pser,ren2023pmil,zhang2025noco}, we use a point-supervised proposal generator to obtain an original pool of candidate proposals $\mathcal P_v=\{(b_j,z_j,\hat c_j)\}_{j=1}^{M_v}$, where $M_v$ is the number of proposals for video $v$. The $j$-th proposal contains an interval $b_j=(s_j,e_j)$, a confidence score $z_j$, and a predicted category $\hat c_j$, where $s_j$ and $e_j$ denote its start and end timestamps and satisfy $s_j<e_j$. Unless explicitly shown, the video index is omitted for brevity, and all point and proposal indices refer to video $v$.

Given the original proposal pool $\mathcal P_v$ and point annotations $\mathcal A_v$, our framework improves pseudo-label quality through two complementary modules: Prior-Guided Initialization (PGI) and Consensus-Guided Refinement (CGR) (Fig.~\ref{fig:framework}(a)). For the $i$-th point annotation in video $v$, PGI uses instance-specific candidate reliability to select an initial pseudo-label $\mathbf y_i^0$ from $\mathcal P_v$. It also retains a candidate subset indexed by $\mathcal P_{v,i}^{\mathrm{ret}}$, together with the candidates' original boundaries and reliability scores. CGR then combines a reliability-weighted consensus of these retained candidates with current detector predictions to construct boundary targets for consistency-controlled pseudo-label refinement.

\subsection{Prior-Guided Initialization (PGI)}
\label{sec:selection}

PGI aims to improve initial pseudo-label quality while retaining alternative candidates for subsequent refinement (Fig.~\ref{fig:framework}(b)). For each annotated point, it assesses candidate reliability using three complementary priors: a model-derived confidence prior and point-derived geometric and conflict priors. The resulting scores guide both initial pseudo-label selection and candidate retention, linking initialization with later refinement.

For the $i$-th annotation $(p_i,c_i)$, we collect from $\mathcal P_v$ the same-class proposals containing $p_i$ and define their index set as:
\begin{equation}
\mathcal P_{v,i}=
\left\{
j\in\{1,\ldots,M_v\}:
\hat c_j=c_i,\;s_j\leq p_i\leq e_j
\right\}.
\label{eq:pool}
\end{equation}

Let $\mathcal I$ denote the indices of annotations with nonempty candidate sets. We initialize and refine pseudo-labels only for $i\in\mathcal I$, excluding annotations with empty candidate sets from detector supervision.

\paragraph{Confidence prior.}
For candidate $j\in\mathcal P_{v,i}$, we define its model-derived confidence prior $\pi_{ij}$ by applying a softmax to the source scores within this set:
\begin{equation}
\pi_{ij}
=
\frac{\exp(z_j)}{\sum_{q\in\mathcal P_{v,i}}\exp(z_q)}.
\label{eq:confidence}
\end{equation}
However, $\pi_{ij}$ is obtained by normalizing CAS-derived scores, which primarily reflect snippet-level action classification confidence and do not fully capture instance-level localization quality. We therefore complement this confidence prior with geometric and conflict priors to assess each candidate's compatibility with the annotated instance.

\paragraph{Geometric prior.}
TSPNet~\cite{xia2024tspnet} has demonstrated the effectiveness of point-derived center cues for proposal confidence estimation. Unlike its learned center-score estimation, PGI directly measures point centrality from the annotated timestamp and proposal boundaries without additional training. For the $j$-th proposal $b_j=(s_j,e_j)$, we define this prior as:
\begin{equation}
\kappa_{ij}
=
\clip_{[0,1]}\!\left(
1-\frac{|p_i-(s_j+e_j)/2|}{(e_j-s_j)/2}
\right),
\label{eq:centrality}
\end{equation}
where $\clip_{[0,1]}(x)=\min\{1,\max\{0,x\}\}$ clips its input to the interval $[0,1]$. This geometric prior favors intervals whose center lies near the annotated point and attenuates those placing it close to a boundary.

\paragraph{Conflict prior.}
A proposal covering additional annotated points can span neighboring instances rather than isolate the target. We collect the other points covered by candidate $j$ as:
\begin{equation}
\mathcal C_{ij}
=\{\ell\in\{1,\ldots,N_v\}:\ell\neq i,\;s_j\leq p_\ell\leq e_j\},
\end{equation}
and measure their accumulated distance from $p_i$:
\begin{equation}
d_{ij}=\sum_{\ell\in\mathcal C_{ij}}|p_\ell-p_i|.
\end{equation}
With timestamps expressed in seconds, we define the conflict prior as:
\begin{equation}
\eta_{ij}
=
\begin{cases}
1, & d_{ij}=0,\\
\tanh\!\left(1/d_{ij}\right), & d_{ij}>0.
\end{cases}
\label{eq:conflict}
\end{equation}
This prior decreases with positive accumulated distance, attenuating intervals that cover additional or more distant annotated instances. It equals one when the candidate covers no other point.

\paragraph{Shared reliability and initial pseudo-labels.}
We multiply the three priors to obtain an instance-specific reliability score. This attenuates candidates with weak geometric or conflict support, even when their classification confidence is high. The highest-scoring candidate provides the initial pseudo-label:
\begin{equation}
\begin{aligned}
r_{ij}&=\pi_{ij}\kappa_{ij}\eta_{ij},\\
j_i^*&=\argmax_{j\in\mathcal P_{v,i}}r_{ij},\\
\mathbf y_i^0&=(s_{j_i^*},e_{j_i^*}).
\end{aligned}
\label{eq:selection}
\end{equation}
The initial pseudo-label inherits the annotated category $c_i$.

\paragraph{Retention of original proposals.}
For each $i\in\mathcal I$, we form the retained candidate index set $\mathcal P_{v,i}^{\mathrm{ret}}\subseteq\mathcal P_{v,i}$ by selecting the $K$ highest-reliability candidates, retaining all candidates when $|\mathcal P_{v,i}|<K$. Their original boundaries $\{b_j\}_{j\in\mathcal P_{v,i}^{\mathrm{ret}}}$ and reliability scores $\{r_{ij}\}_{j\in\mathcal P_{v,i}^{\mathrm{ret}}}$ remain fixed throughout training.

\subsection{Consensus-Guided Refinement (CGR)}
\label{sec:cgr}
CGR refines pseudo-labels using retained proposal evidence and current detector predictions (Fig.~\ref{fig:framework}(c)). At each round, it combines consensus from retained proposals with the detector prediction to construct a boundary target, then applies a momentum update under a progressively relaxed consistency criterion.

\paragraph{Current detector predictions.}
At round $k$, we apply the PGI reliability rule to detector predictions covering $p_i$ and select the highest-reliability interval $b_i^{\mathrm{det},k}$. The annotated category $c_i$ remains the class supervision.

\paragraph{Consensus from retained original proposals.}
We assign each retained candidate a weight based on its fixed reliability and temporal agreement with $b_i^{\mathrm{det},k}$, normalize these weights, and aggregate the candidates' start and end timestamps:

\begin{equation}
\begin{aligned}
w_{ij}^{k}
&=r_{ij}\exp\!\left(\tiou(b_j,b_i^{\mathrm{det},k})\right),\\
\bar w_{ij}^{k}
&=\frac{w_{ij}^{k}}{\sum_{q\in\mathcal P_{v,i}^{\mathrm{ret}}}w_{iq}^{k}},\\
b_i^{\mathrm{con},k}
&=\sum_{j\in\mathcal P_{v,i}^{\mathrm{ret}}}\bar w_{ij}^{k}b_j,
\end{aligned}
\label{eq:consensus}
\end{equation}
where $b_i^{\mathrm{con},k}$ is the consensus interval for point $i$ at round $k$. The normalized weights favor reliable candidates agreeing with the detector. Although $b_j$ and $r_{ij}$ remain fixed, the consensus is recomputed from current predictions at each update, allowing retained alternatives to continually support refinement.

\paragraph{Boundary target construction.}
A binary gate $g_i^k$ determines whether proposal consensus is fused with the detector prediction to construct the boundary target:

\begin{equation}
b_i^{\mathrm{tar},k}
=b_i^{\mathrm{det},k}
+\tfrac12 g_i^k
\left(b_i^{\mathrm{con},k}-b_i^{\mathrm{det},k}\right),
\label{eq:target}
\end{equation}
where $b_i^{\mathrm{tar},k}$ is the target interval for point $i$ at round $k$. It equally weights the two intervals when $g_i^k=1$ and uses the detector prediction alone otherwise. We define the gate by:

\begin{equation}
g_i^k=\mathbf1\!\left[
\tiou(b_i^{\mathrm{det},k},b_i^{\mathrm{con},k})\geq\tau_{\mathrm{iou}}
\right],
\label{eq:gate}
\end{equation}
where $\mathbf1[\cdot]$ is the indicator function and $\tau_{\mathrm{iou}}\in[0,1]$ is the minimum detector--consensus tIoU required for fusion.

\paragraph{Consistency-controlled pseudo-label refinement.}
Consistency with the current pseudo-label controls update eligibility. For each $i\in\mathcal I$, pseudo-label $\mathbf y_i^k$ is eligible for refinement at round $k$ if $b_i^{\mathrm{det},k}$ exists and $\tiou(b_i^{\mathrm{det},k},\mathbf y_i^k)\geq\tau_{\mathrm{con}}^k$, where $\tau_{\mathrm{con}}^k\in[0,1]$ specifies the required consistency. Let $\mathcal I_k\subseteq\mathcal I$ denote the indices satisfying these conditions. We refine the eligible pseudo-labels toward their boundary targets using:

\begin{equation}
\mathbf y_i^{k+1}
=(1-\mu)\mathbf y_i^k+\mu b_i^{\mathrm{tar},k},
\qquad i\in\mathcal I_k,
\label{eq:update}
\end{equation}
where $\mu\in(0,1]$ controls the target's contribution. The remaining pseudo-labels and all annotated categories remain unchanged. The refined pseudo-labels supervise the next training period, and CGR repeats target construction and refinement at subsequent rounds.

\begin{table*}[t]
\centering
\caption{Comparison with state-of-the-art methods on THUMOS14. Results are reported as mAP (\%) at different tIoU thresholds. Frame-level, video-level, and point-level supervision correspond to fully supervised TAL, WTAL, and PTAL, respectively. Bold and underlined values indicate the best and second-best point-supervised results, respectively. Ours$^{\dagger}$ uses the same HR-Pro candidate proposal pool as PseR.}
\label{tab:main_thumos}
\resizebox{0.94\textwidth}{!}{%
\begin{tabular}{c|l|ccccccc|ccc}
\toprule
\multirow{2}{*}{\textbf{Supervision}} & \multicolumn{1}{c|}{\multirow{2}{*}{\textbf{Method}}} & \multicolumn{7}{c|}{\textbf{mAP@IoU(\%)}} & \multirow{2}{*}{\begin{tabular}[c]{@{}c@{}}\textbf{AVG}\\ \textbf{(0.1:0.5)}\end{tabular}} & \multirow{2}{*}{\begin{tabular}[c]{@{}c@{}}\textbf{AVG}\\ \textbf{(0.3:0.7)}\end{tabular}} & \multirow{2}{*}{\begin{tabular}[c]{@{}c@{}}\textbf{AVG}\\ \textbf{(0.1:0.7)}\end{tabular}} \\
\cmidrule{3-9}
 & \multicolumn{1}{c|}{} & \textbf{0.1} & \textbf{0.2} & \textbf{0.3} & \textbf{0.4} & \textbf{0.5} & \textbf{0.6} & \textbf{0.7} &  &  &  \\ \midrule
\multirow{3}{*}{Frame-level} 
 & ActionFormer \textsubscript{\textcolor{gray}{ECCV'22}}~\cite{zhang2022actionformer} & - & - & 82.1 & 77.8 & 71.0 & 59.4 & 43.9 & - & 66.8 & - \\
 & TriDet \textsubscript{\textcolor{gray}{CVPR'23}}~\cite{shi2023tridet} & - & - & 83.6 & 80.1 & 72.9 & 62.4 & 47.4 & - & 69.3 & - \\ 
 & MDVLM-TAL \textsubscript{\textcolor{gray}{NeurIPS'26}}~\cite{wang2026mdvlmtal} & - & - & 84.6 & 80.8 & 74.4 & 63.5 & 60.1 & - & 72.7 & - \\ \midrule
\multirow{7}{*}{Video-level} & P-MIL \textsubscript{\textcolor{gray}{CVPR'23}}~\cite{ren2023pmil} & 71.8 & 67.5 & 58.9 & 49.0 & 40.0 & 27.1 & 15.1 & 57.4 & 38.0 & 47.0 \\
 & PivoTAL \textsubscript{\textcolor{gray}{CVPR'23}}~\cite{rizve2023pivotal} & 74.1 & 69.6 & 61.7 & 52.1 & 42.8 & 30.6 & 16.7 & 60.1 & 40.8 & 49.6 \\
 & ISSF \textsubscript{\textcolor{gray}{AAAI'24}}~\cite{yun2023weakly} & 72.4 & 66.9 & 58.4 & 49.7 & 41.8 & 25.5 & 12.8 & 57.8 & 37.6 & 46.8 \\ 
 & PVLR \textsubscript{\textcolor{gray}{MM'24}}~\cite{lim2024pvlr} & 74.9 & 69.9 & 61.4 & 53.1 & 45.1 & 30.5 & 17.1 & 60.9 & 41.4 & 50.3 \\
 & PseudoFormer \textsubscript{\textcolor{gray}{CVPR'25}}~\cite{liu2025pseudoformer} & 76.8 & 72.6 & 65.4 & 55.6 & 44.1 & 31.3 & 17.1 & 62.9 & 42.7 & 51.9 \\ 
 & CRC \textsubscript{\textcolor{gray}{TPAMI'25}}~\cite{yang2025crc} & 72.9 & 68.0 & 59.8 & 51.3 & 42.7 & 27.8 & 15.1 & 58.9 & 39.3 & 48.2 \\ 
 & ACL-Net \textsubscript{\textcolor{gray}{TIP'26}}~\cite{li2026aclnet} & 81.0 & 76.9 & 70.0 & 60.4 & 51.3 & 36.8 & 21.5 & 67.9 & 48.0 & 56.8 \\ 
 \midrule
\multirow{17}{*}{Point-level} & 
SF-Net \textsubscript{\textcolor{gray}{ECCV'20}}~\cite{ma2020sfnet} & 68.3 & 62.3 & 52.8 & 42.2 & 30.5 & 20.6 & 12.0 & 51.2 & 31.6 & 41.2 \\
 & Ju \textit{et al.} \textsubscript{\textcolor{gray}{ICCV'21}}~\cite{ju2021divide} & 72.3 & 64.7 & 58.2 & 47.1 & 35.9 & 23.0 & 12.8 & 55.6 & 35.4 & 44.9 \\
 & LACP \textsubscript{\textcolor{gray}{ICCV'21}}~\cite{lee2021lacp} & 75.7 & 71.4 & 64.6 & 56.5 & 45.3 & 34.5 & 21.8 & 62.7 & 44.5 & 52.8 \\
 & CRRC-Net \textsubscript{\textcolor{gray}{TIP'22}}~\cite{fu2022crrcnet} & 77.8 & 73.5 & 67.1 & 57.9 & 46.6 & 33.7 & 19.8 & 64.6 & 45.1 & 53.8 \\
 & LPR \textsubscript{\textcolor{gray}{MMSJ'23}}~\cite{fang2023lpr} & 85.3 & 82.5 & 76.6 & 67.8 & 55.3 & 38.5 & 21.7 & 73.5 & 52.0 & 61.1 \\
 & Yin \textit{et al.} \textsubscript{\textcolor{gray}{BMVC'23}}~\cite{yin2023proposal} & 77.1 & 72.6 & 65.9 & 54.4 & 44.9 & 33.1 & 20.2 & 63.3 & 43.9 & 52.6 \\
 & HR-Pro \textsubscript{\textcolor{gray}{AAAI'24}}~\cite{zhang2023hrpro} & 85.6 & 81.6 & 74.3 & 64.3 & 52.2 & 39.8 & 24.8 & 71.6 & 51.1 & 60.3 \\
 & TSPNet \textsubscript{\textcolor{gray}{CVPR'24}}~\cite{xia2024tspnet} & 82.3 & 77.6 & 70.1 & 60.0 & 49.4 & 37.6 & 24.5 & 67.9 & 48.3 & 57.4 \\
 & SMBD \textsubscript{\textcolor{gray}{ECCV'24}}~\cite{liu2025smbd} & - & - & 66.0 & 57.9 & 47.0 & 36.0 & 22.0 & 64.2 & 45.7 & - \\
 & SNPR \textsubscript{\textcolor{gray}{TIP'24}}~\cite{li2024snpr} & 77.9 & 73.9 & 66.6 & 59.4 & 48.6 & 36.7 & 22.7 & 65.3 & 46.8 & 55.1 \\ 
 & AAPL \textsubscript{\textcolor{gray}{AAAI'25}}~\cite{yoshida2025aapl} & 64.3 & - & 54.6 & - & 35.2 & - & 14.0 & - & - & 42.8 \\ 
 & NoCo \textsubscript{\textcolor{gray}{AAAI'25}}~\cite{zhang2025noco} & 86.0 & 82.1 & 75.6 & 65.7 & 53.6 & 40.6 & 25.3 & 72.6 & 52.2 & 61.3 \\ 
 & QROT \textsubscript{\textcolor{gray}{CVPR'25}}~\cite{liu2025qrot} & - & - & 73.1 & 64.4 & 54.3 & 41.3 & 27.4 & 72.3 & 52.1 & - \\ 
 & TRA \textsubscript{\textcolor{gray}{ESWA'26}}~\cite{ma2027tra} & \textbf{86.6} & \underline{83.6} & 76.3 & 66.7 & 55.2 & 41.4 & 26.5 & 73.7 & 53.2 & 62.3 \\ 
 & PseR \textsubscript{\textcolor{gray}{TMM'26}}~\cite{fang2026pser} & 85.5 & 83.2 & \underline{78.7} & 70.9 & 60.1 & 45.4 & 29.6 & 75.7 & 56.9 & 64.8 \\
 & Ours$^{\dagger}$ & \underline{86.3} & 83.0 & 78.6 & \underline{73.4} & \underline{63.1} & \underline{50.9} & \underline{33.8} & \underline{76.9} & \underline{60.0} & \underline{67.0} \\
 & Ours & \textbf{86.6} & \textbf{83.8} & \textbf{80.4} & \textbf{74.8} & \textbf{65.5} & \textbf{52.6} & \textbf{35.5} & \textbf{78.2} & \textbf{61.8} & \textbf{68.5} \\ \bottomrule
\end{tabular}
}
\end{table*}

\subsection{Training and Inference}
\label{sec:training}

\paragraph{Training.}
We train the detector using the initial pseudo-labels from PGI and the annotated categories. After each CGR update, the refined pseudo-labels provide temporal supervision for subsequent training.

\paragraph{Inference.}
During inference, the trained detector directly predicts action categories, temporal boundaries, and confidence scores for an unseen video. PGI, CGR, point annotations, and the original candidate pool are used only during training.
\section{Experiments}
\label{sec:experiment}

\subsection{Experimental Setup}
\label{sec:experimental_setup}
\paragraph{Datasets and metrics.}
We evaluate on THUMOS14~\cite{idrees2017thumos}, ActivityNet1.3~\cite{caba2015activitynet}, GTEA~\cite{fathi2011gtea}, and BEOID~\cite{damen2014beoid}, using the point annotations and evaluation splits of HR-Pro~\cite{zhang2023hrpro}. Detection performance is measured by mAP at multiple tIoU thresholds. AVG averages thresholds 0.1--0.7 on THUMOS14, GTEA, and BEOID, and 0.5--0.95 on ActivityNet1.3. Pseudo-label quality is measured by mean tIoU with official GT and the fraction with tIoU $\geq0.7$.

\paragraph{Implementation.}
Following existing methods~\cite{zhang2023hrpro,fang2026pser}, we extract I3D features from 16-frame snippets using a network pretrained on Kinetics-400~\cite{carreira2017kinetics}. HR-Pro~\cite{zhang2023hrpro} generates the candidate proposal pool $\mathcal P_v$ from these features. We train ActionFormer~\cite{zhang2022actionformer} with AdamW and a weight decay of $0.05$. The initial learning rate is $10^{-3}$ on ActivityNet1.3 and $10^{-4}$ on the other datasets. For THUMOS14, ActivityNet1.3, GTEA, and BEOID, we use batch sizes of $2$, $16$, $1$, and $2$ and train for $35$, $10$, $60$, and $50$ epochs, respectively.

Across all datasets, PGI retains up to $K=10$ candidates per point, and CGR uses $\tau_{\mathrm{iou}}=0.5$ for consensus gating and $\mu=0.9$ for momentum updates. We schedule CGR updates between approximately $20\%$ and $80\%$ of training. In the default schedule, the consistency threshold decays linearly from $1$ to $0$:
\begin{equation}
\tau_{\mathrm{con}}^k=\frac{e_{\mathrm{last}}-e_k}{e_{\mathrm{last}}-e_{\mathrm{first}}},
\label{eq:consistency_schedule}
\end{equation}
where $e_k$ is the epoch of update $k$, and $e_{\mathrm{first}}$ and $e_{\mathrm{last}}$ are the first and last update epochs. At inference, we apply Soft-NMS to suppress redundant predictions.

\subsection{Comparison with State-of-the-art Methods}
\label{sec:comparisonsota}

\paragraph{Results on THUMOS14.}
As shown in Table~\ref{tab:main_thumos}, our default configuration achieves the highest average mAP among point-supervised methods across all three tIoU ranges on THUMOS14. Our method achieves 68.5\% average mAP over tIoU thresholds 0.1--0.7, surpassing PseR by 3.7 percentage points. At tIoU thresholds of 0.5 and 0.7, the gains are 5.4 and 5.9 percentage points, respectively. Using the HR-Pro candidate proposal pool provided by PseR, Ours$^{\dagger}$ achieves 67.0\% average mAP over tIoU thresholds 0.1--0.7, improving on PseR by 2.2 percentage points. These results demonstrate the effectiveness of our framework for PTAL on THUMOS14.

\paragraph{Results on ActivityNet1.3.}
As shown in Table~\ref{tab:main_anet}, our method achieves 30.9\% average mAP over tIoU thresholds 0.5--0.95, the highest among the compared methods and 0.3 percentage points above PseR. At tIoU 0.75, it obtains 30.7\% mAP, improving over PseR's 29.6\% by 1.1 percentage points. These results show that our framework also improves overall detection performance on ActivityNet1.3.

\begin{table}[!t]
\centering
\caption{Comparison with state-of-the-art methods on ActivityNet1.3. Results are reported as mAP (\%) at different tIoU thresholds.}
\label{tab:main_anet}
\resizebox{0.95\columnwidth}{!}{%
\begin{tabular}{c|l|cccc}
\toprule
\multirow{2}{*}{\textbf{Supervision}} & \multicolumn{1}{c|}{\multirow{2}{*}{\textbf{Method}}} & \multicolumn{4}{c}{\textbf{mAP@tIoU (\%)}} \\
\cmidrule{3-6}
 & \multicolumn{1}{c|}{} & \textbf{0.5} & \textbf{0.75} & \textbf{0.95} & \textbf{AVG{[}0.5:0.95{]}} \\ \midrule
\multirow{6}{*}{Video-level} & PivoTAL \textsubscript{\textcolor{gray}{CVPR'23}} & 45.1 & 28.2 & 5.0 & 28.1 \\
 & ISSF \textsubscript{\textcolor{gray}{AAAI'24}} & 39.4 & 25.8 & 6.4 & 25.8 \\
 & PVLR \textsubscript{\textcolor{gray}{MM'24}} & 43.6 & 27.4 & 6.5 & 27.4 \\
 & PseudoFormer \textsubscript{\textcolor{gray}{CVPR'25}} & 46.9 & 28.6 & 6.5 & 29.0 \\ 
 & CRC \textsubscript{\textcolor{gray}{TPAMI'25}} & 42.0 & 26.2 & 5.4 & 26.2 \\ 
 & ACL-Net \textsubscript{\textcolor{gray}{TIP'26}} & 43.4 & 27.6 & 6.6 & 27.6 \\ 
\midrule
\multirow{8}{*}{Point-level} & LACP \textsubscript{\textcolor{gray}{ICCV'21}} & 40.4 & 24.6 & 5.7 & 25.1 \\
 & CRRC-Net \textsubscript{\textcolor{gray}{TIP'22}} & 39.8 & 24.1 & 5.9 & 24.0 \\
 & Yin \textit{et al.} \textsubscript{\textcolor{gray}{BMVC'23}} & 48.3 & 27.8 & 7.0 & 29.1 \\
 & HR-Pro \textsubscript{\textcolor{gray}{AAAI'24}} & 42.8 & 27.2 & \textbf{8.0} & 27.1 \\
 & AAPL \textsubscript{\textcolor{gray}{AAAI'25}} & 39.6 & 24.3 & 5.6 & 24.7 \\ 
 & TRA \textsubscript{\textcolor{gray}{ESWA'26}} & 45.3 & 27.4 & 6.8 & 28.0 \\ 
 & PseR \textsubscript{\textcolor{gray}{TMM'26}} & \textbf{49.5} & \underline{29.6} & \underline{7.8} & \underline{30.6} \\
 & Ours & \underline{49.0} & \textbf{30.7} & 7.2 & \textbf{30.9} \\ \bottomrule
\end{tabular}

}
\end{table}

\paragraph{Results on GTEA and BEOID.}
As shown in Table~\ref{tab:small_datasets}, our method achieves 62.2\% and 74.2\% AVG mAP on GTEA and BEOID, respectively. It surpasses TRA on GTEA by 8.1 percentage points and PseR on BEOID by 5.3 percentage points. Our method also outperforms the compared methods at every reported tIoU threshold on both datasets. In particular, at tIoU 0.7 on BEOID, it achieves 42.3\% mAP, exceeding TRA's 31.4\% by 10.9 percentage points. The complete per-threshold comparisons are provided in the supplementary material.

\begin{table}[!t]
\centering
\caption{Comparison with point-supervised methods on GTEA and BEOID. Results are AVG mAP (\%) over tIoU thresholds 0.1--0.7; per-threshold results appear in the supplementary material.}
\label{tab:small_datasets}
\small
\setlength{\tabcolsep}{5pt}
\begin{tabular}{lcc}
\toprule
\textbf{Method} & \textbf{GTEA} & \textbf{BEOID} \\
\midrule
SF-Net \textsubscript{\textcolor{gray}{ECCV'20}} & 31.0 & 30.9 \\
Ju \textit{et al.} \textsubscript{\textcolor{gray}{ICCV'21}} & 33.7 & 34.9 \\
LACP \textsubscript{\textcolor{gray}{ICCV'21}} & 43.5 & 51.8 \\
Yin \textit{et al.} \textsubscript{\textcolor{gray}{BMVC'23}} & 45.1 & 54.2 \\
HR-Pro \textsubscript{\textcolor{gray}{AAAI'24}} & 47.3 & 59.4 \\
TSPNet \textsubscript{\textcolor{gray}{CVPR'24}} & 49.0 & 59.6 \\
SMBD \textsubscript{\textcolor{gray}{ECCV'24}} & 47.4 & 57.4 \\
SNPR \textsubscript{\textcolor{gray}{TIP'24}} & 46.6 & 53.1 \\
AAPL \textsubscript{\textcolor{gray}{AAAI'25}} & 46.3 & 55.2 \\
TRA \textsubscript{\textcolor{gray}{ESWA'26}} & \underline{54.1} & 63.7 \\
PseR \textsubscript{\textcolor{gray}{TMM'26}} & 52.3 & \underline{68.9} \\
Ours & \textbf{62.2} & \textbf{74.2} \\
\bottomrule
\end{tabular}

\end{table}

\begin{figure*}[t]
\centering
\includegraphics[width=0.75\textwidth]{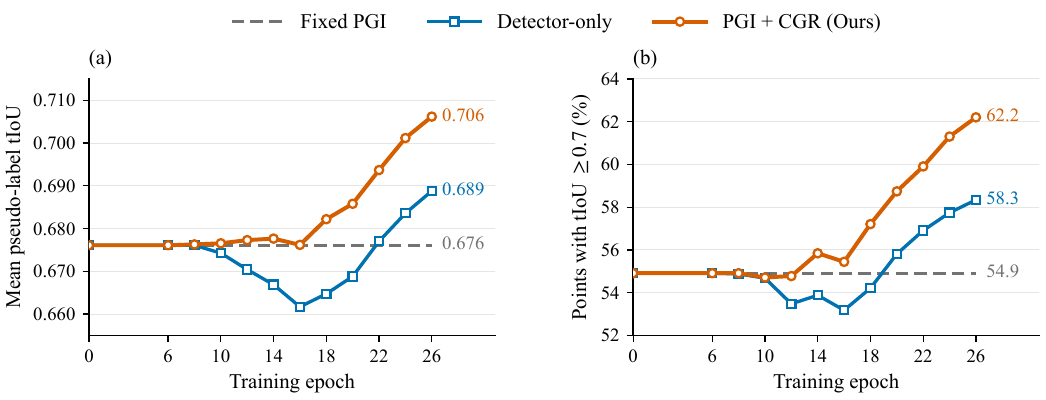}
\caption{\textbf{Pseudo-label refinement dynamics on THUMOS14.} (a) Mean pseudo-label tIoU. (b) Fraction of points with tIoU $\geq 0.7$. All methods share the same PGI initialization and are evaluated against official GT over all training points, assigning zero tIoU to points without a pseudo-label. Epoch 0 denotes initialization; markers indicate refinement checkpoints. Final values are annotated.}
\label{fig:refinement_dynamics}
\end{figure*}

\subsection{Ablation Studies}
\label{sec:ablation}
In this subsection, we conduct a series of ablation studies to evaluate the contributions of the proposed components.

\paragraph{Effectiveness of PGI and CGR.}
Table~\ref{tab:ablation_modules} compares training with fixed HR-Pro pseudo-labels, fixed PGI pseudo-labels, and PGI followed by CGR. PGI improves AVG mAP by 0.5--2.9 percentage points across the four benchmarks, while CGR adds a further 0.9--5.2 percentage points. CGR is particularly effective on GTEA, adding 5.2 percentage points after PGI's 0.5-point gain.

\begin{table}[!t]
\centering
\caption{Component ablation of PGI and CGR. Results are average mAP (\%) over tIoU thresholds 0.1--0.7 for THUMOS14, GTEA, and BEOID, and 0.5--0.95 for ActivityNet1.3. A checkmark indicates an enabled module.}
\label{tab:ablation_modules}
\small
\setlength{\tabcolsep}{3pt}
\resizebox{0.83\columnwidth}{!}{%
\begin{tabular}{cccccc}
\toprule
\textbf{PGI} & \textbf{CGR} & \textbf{THUMOS14} & \textbf{ActivityNet1.3} & \textbf{GTEA} & \textbf{BEOID} \\
\midrule
-- & -- & 64.1 & 29.1 & 56.5 & 69.5 \\
$\checkmark$ & -- & 67.0 & 30.0 & 57.0 & 72.2 \\
\rowcolor{orange!12}
$\checkmark$ & $\checkmark$ & 68.5 & 30.9 & 62.2 & 74.2 \\
\bottomrule
\end{tabular}

}
\end{table}

\paragraph{Point-derived priors.}
We vary the geometric and conflict priors while keeping the confidence prior and original candidate pool fixed. As shown in Table~\ref{tab:ablation_priors}, the geometric prior yields a higher fraction of points with tIoU $\geq 0.7$ (50.7\% vs. 46.8\%), while the conflict prior yields higher mean tIoU (0.649 vs. 0.599). Combining both improves mean tIoU from 0.569 to 0.676 and the high-quality fraction from 39.8\% to 54.9\% over confidence alone.

\begin{table}[!t]
\centering
\caption{Ablation of PGI priors on THUMOS14. Both metrics include all training points. Points without a valid candidate are assigned zero tIoU.}
\label{tab:ablation_priors}
\small
\setlength{\tabcolsep}{3pt}
\resizebox{0.78\columnwidth}{!}{%
\begin{tabular}{ccccc}
\toprule
\multirow{2}{*}{\textbf{Confidence}} & \multirow{2}{*}{\textbf{Geometric}} & \multirow{2}{*}{\textbf{Conflict}} & \textbf{Mean} & \textbf{\boldmath tIoU $\geq 0.7$} \\
 & & & \textbf{tIoU} & \textbf{(\%)} \\
\midrule
$\checkmark$ & -- & -- & 0.569 & 39.8 \\ 
$\checkmark$ & $\checkmark$ & -- & 0.599 & 50.7 \\ 
$\checkmark$ & -- & $\checkmark$ & 0.649 & 46.8 \\ 
\rowcolor{orange!12}
$\checkmark$ & $\checkmark$ & $\checkmark$ & 0.676 & 54.9 \\ 
\bottomrule
\end{tabular}

}
\end{table}

\begin{figure*}[t]
\centering
\includegraphics[width=0.95\textwidth]{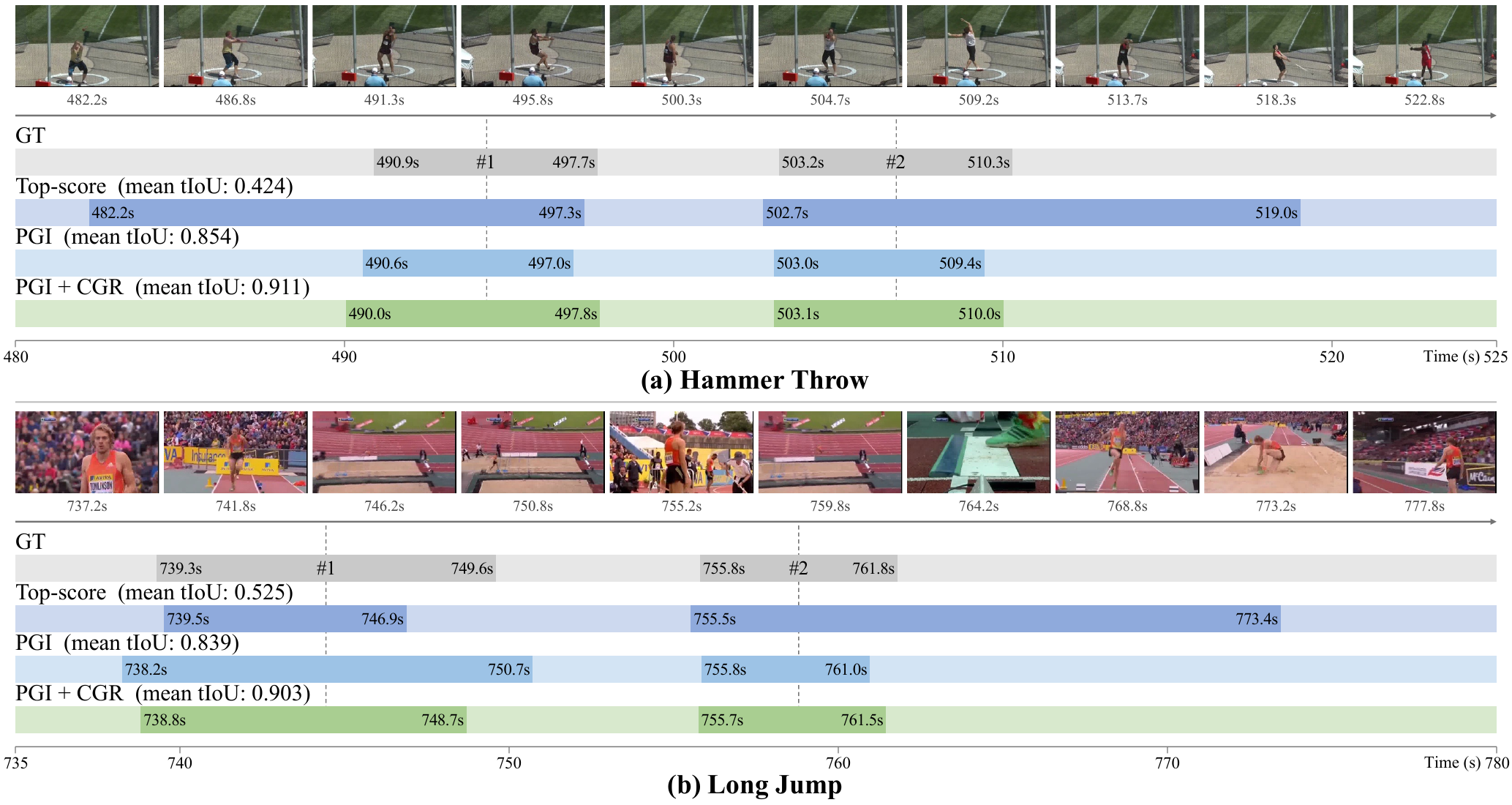}
\caption{\textbf{Pseudo-labels on THUMOS14 training clips.} Top-score selection, PGI, and final CGR labels are compared with GT for HammerThrow and LongJump. Dashed lines mark annotated points. Row labels report mean tIoU over the two displayed instances.}
\label{fig:pseudo_label_qualitative}
\end{figure*}

\paragraph{Retained proposal evidence.}
With the same PGI initialization, we compare detector-only refinement (using detector predictions as boundary targets), guidance from the initial boundary ($K=1$), and full CGR ($K=10$) on THUMOS14. In Table~\ref{tab:proposal_consensus}, CGR improves AVG mAP over the first two variants by 1.1 and 0.5 percentage points, respectively, and achieves the best pseudo-label quality on both metrics. Relative to initial-boundary guidance, CGR also raises mean pseudo-label tIoU from 0.696 to 0.706 and the high-quality fraction from 61.3\% to 62.2\%. These results demonstrate the benefit of aggregating alternative proposal boundaries during refinement.

\begin{table}[!t]
\centering
\caption{Contribution of proposal consensus on THUMOS14. AVG mAP (\%) is computed over tIoU thresholds 0.1--0.7 at the final training epoch. Both quality metrics compare final pseudo-labels with official GT over all training points, with zero tIoU for points without a pseudo-label. The highlighted row denotes full CGR.}
\label{tab:proposal_consensus}
\small
\setlength{\tabcolsep}{3pt}
\resizebox{0.85\columnwidth}{!}{%
\begin{tabular}{lccc}
\toprule
\multicolumn{1}{c}{\multirow{2}{*}{\textbf{Boundary evidence}}} & \textbf{AVG mAP} & \textbf{Mean} & \textbf{\boldmath tIoU $\geq 0.7$} \\
 & \textbf{(\%)} & \textbf{tIoU} & \textbf{(\%)} \\
\midrule
Detector-only & 67.4 & 0.689 & 58.3 \\
Initial boundary & 68.0 & 0.696 & 61.3 \\
\rowcolor{orange!12}
Proposal consensus & 68.5 & 0.706 & 62.2 \\
\bottomrule
\end{tabular}

}
\end{table}

\paragraph{Refinement dynamics.}
Figure~\ref{fig:refinement_dynamics} shows that detector-only refinement declines during intermediate training before recovering, whereas CGR remains relatively stable and improves at later updates. Over all training points, CGR increases mean pseudo-label tIoU from 0.676 to 0.706 and the fraction with tIoU $\geq 0.7$ from 54.9\% to 62.2\%, a gain of 7.3 percentage points over fixed PGI supervision.

\paragraph{Additional analyses.}
The supplementary material reports consistency and hyperparameter ablations, along with evaluations using LACP~\cite{lee2021lacp} and HR-Pro~\cite{zhang2023hrpro} proposals with ActionFormer~\cite{zhang2022actionformer} and TriDet~\cite{shi2023tridet}, where our method improves AVG mAP over tIoU thresholds 0.3--0.7 by 4.6--5.4 percentage points.

\subsection{Qualitative Analysis}
\label{sec:qualitative}
Figure~\ref{fig:pseudo_label_qualitative} illustrates how PGI and CGR improve localization on two THUMOS14 training clips. In Hammer Throw, top-score proposals extend into neighboring action instances. In Long Jump, one proposal ends before the action finishes, while the other extends well beyond its ground-truth boundary. PGI selects better-aligned intervals, raising mean tIoU from 0.424 to 0.854 for Hammer Throw and from 0.525 to 0.839 for Long Jump. CGR further adjusts the boundaries, reaching 0.911 and 0.903, respectively. Detection visualizations appear in the supplementary material.
\section{Conclusion}
\label{sec:conclusion}

We presented a pseudo-label learning framework for point-supervised temporal action localization that reuses original proposal evidence across initialization and refinement. PGI combines confidence, geometric, and conflict priors to select initial pseudo-labels and retain reliable alternatives. CGR combines their reliability-weighted consensus with detector predictions to construct boundary targets for consistency-controlled momentum updates. Experiments on four benchmarks demonstrate competitive detection performance, while ablations show how initialization and retained proposal evidence improve pseudo-label quality. Further evaluations demonstrate consistent gains across proposal generators and detectors.

{
    \small
    \bibliographystyle{ieeenat_fullname}
    \bibliography{main}
}



\clearpage

\setcounter{section}{0}
\setcounter{table}{0}
\setcounter{figure}{0}
\setcounter{equation}{0}

\renewcommand{\thesection}{\Alph{section}}
\renewcommand{\thetable}{S\arabic{table}}
\renewcommand{\thefigure}{S\arabic{figure}}
\renewcommand{\theequation}{S\arabic{equation}}

\twocolumn[
    \begin{center}
        \Large\bfseries Supplementary Material
    \end{center}
    \vspace{0.5em}
]

\section{Additional Experimental Details}

\label{supp:setup}

\paragraph{Datasets.}
We evaluate our framework on four benchmarks using the point annotations provided by HR-Pro~\cite{zhang2023hrpro}. The dataset sizes and evaluation splits are summarized below.

\begin{itemize}
    \item \textbf{THUMOS14}~\cite{idrees2017thumos} contains 413 untrimmed videos spanning 20 action categories. We use 200 videos for training and the remaining 213 for testing.
    \item \textbf{ActivityNet1.3}~\cite{caba2015activitynet} comprises 10,024 training, 4,926 validation, and 5,044 test videos. Following the standard protocol~\cite{zhang2023hrpro,fang2026pser}, we train on the training split and evaluate on the validation split.
    \item \textbf{GTEA}~\cite{fathi2011gtea} consists of 28 videos depicting seven fine-grained daily activities in a kitchen, with approximately 1,800 frames per video.
    \item \textbf{BEOID}~\cite{damen2014beoid} contains 58 videos covering 30 action categories, with an average of 12.5 action instances per video. We follow the training and testing split used in~\cite{moltisanti2019action}.
\end{itemize}

\paragraph{Evaluation metric.}
We report mean average precision (mAP) at multiple temporal intersection-over-union (tIoU) thresholds. Unless otherwise specified, AVG mAP is the mean of the mAP values across thresholds from 0.1 to 0.7 in steps of 0.1 for THUMOS14, GTEA, and BEOID, and from 0.50 to 0.95 in steps of 0.05 for ActivityNet1.3. We assess pseudo-label quality over all annotated action instances using mean tIoU against official ground truth (GT) and the fraction of instances with tIoU $\geq 0.7$. Instances without a corresponding pseudo-label are assigned a tIoU of zero.

\section{Extended Benchmark Comparisons}
\label{supp:comparisons}
Tables~\ref{supp:main_gtea} and~\ref{supp:main_beoid} provide per-threshold comparisons on GTEA and BEOID, complementing the AVG mAP results in the main paper. Our framework outperforms the compared methods at every reported tIoU threshold on both datasets. At tIoU 0.7, it achieves 25.5\% mAP on GTEA and 42.3\% on BEOID, exceeding the strongest compared results by 2.1 and 10.9 percentage points, respectively. These gains extend the average-performance improvements to stricter localization criteria.

\begin{table}[t]
\centering
\caption{Comparison with point-supervised methods on GTEA. Results are reported as mAP (\%) at different tIoU thresholds.}
\label{supp:main_gtea}
\resizebox{\columnwidth}{!}{%
\begin{tabular}{l|ccccc}
\toprule
\multicolumn{1}{c|}{\multirow{2}{*}{\textbf{Method}}} & \multicolumn{5}{c}{\textbf{mAP@tIoU (\%)}} \\ \cmidrule{2-6}
\multicolumn{1}{c|}{} & \textbf{0.1} & \textbf{0.3} & \textbf{0.5} & \textbf{0.7} & \textbf{AVG{[}0.1:0.7{]}} \\ \midrule
SF-Net \textsubscript{\textcolor{gray}{ECCV'20}} & 58.0 & 37.9 & 19.3 & 11.9 & 31.0 \\
Ju \textit{et al.} \textsubscript{\textcolor{gray}{ICCV'21}} & 59.7 & 38.3 & 21.9 & 18.1 & 33.7 \\
LACP \textsubscript{\textcolor{gray}{ICCV'21}} & 63.9 & 55.7 & 33.9 & 20.8 & 43.5 \\
Yin \textit{et al.} \textsubscript{\textcolor{gray}{BMVC'23}} & 63.1 & 52.1 & 37.3 & 22.2 & 45.1 \\
HR-Pro \textsubscript{\textcolor{gray}{AAAI'24}} & 72.6 & 61.1 & 37.3 & 17.5 & 47.3 \\
TSPNet \textsubscript{\textcolor{gray}{CVPR'24}} & 74.6 & 60.9 & 39.5 & 16.6 & 49.0 \\
SMBD \textsubscript{\textcolor{gray}{ECCV'24}} & 75.0 & 61.3 & 41.1 & 14.2 & 47.4 \\
SNPR \textsubscript{\textcolor{gray}{TIP'24}} & 74.3 & 62.8 & 35.7 & 13.7 & 46.6 \\
AAPL \textsubscript{\textcolor{gray}{AAAI'25}} & 70.3 & 54.4 & 37.7 & \underline{23.4} & 46.3 \\
TRA \textsubscript{\textcolor{gray}{ESWA'26}} & \underline{77.9} & \underline{67.8} & \underline{47.2} & 22.4 & \underline{54.1} \\
PseR \textsubscript{\textcolor{gray}{TMM'26}} & 72.7 & 67.7 & 42.9 & 19.9 & 52.3 \\
Ours & \textbf{84.2} & \textbf{77.7} & \textbf{57.9} & \textbf{25.5} & \textbf{62.2} \\ \bottomrule
\end{tabular}

}
\end{table}

\begin{table}[t]
\centering
\caption{Comparison with point-supervised methods on BEOID. Results are reported as mAP (\%) at different tIoU thresholds.}
\label{supp:main_beoid}
\resizebox{\columnwidth}{!}{%
\begin{tabular}{l|ccccc}
\toprule
\multicolumn{1}{c|}{\multirow{2}{*}{\textbf{Method}}} & \multicolumn{5}{c}{\textbf{mAP@tIoU (\%)}} \\ \cmidrule{2-6}
\multicolumn{1}{c|}{} & \textbf{0.1} & \textbf{0.3} & \textbf{0.5} & \textbf{0.7} & \textbf{AVG{[}0.1:0.7{]}} \\ \midrule
SF-Net \textsubscript{\textcolor{gray}{ECCV'20}} & 62.9 & 40.6 & 16.7 & 3.5 & 30.9 \\
Ju \textit{et al.} \textsubscript{\textcolor{gray}{ICCV'21}} & 63.2 & 46.8 & 20.9 & 5.8 & 34.9 \\
LACP \textsubscript{\textcolor{gray}{ICCV'21}} & 76.9 & 61.4 & 42.7 & 25.1 & 51.8 \\
Yin \textit{et al.} \textsubscript{\textcolor{gray}{BMVC'23}} & 78.2 & 65.3 & 45.1 & 26.6 & 54.2 \\
HR-Pro \textsubscript{\textcolor{gray}{AAAI'24}} & 78.5 & 72.1 & 55.3 & 26.1 & 59.4 \\
TSPNet \textsubscript{\textcolor{gray}{CVPR'24}} & 83.8 & 73.0 & 51.1 & 23.8 & 59.6 \\
SMBD \textsubscript{\textcolor{gray}{ECCV'24}} & 78.2 & 71.0 & 52.5 & 25.2 & 57.4 \\
SNPR \textsubscript{\textcolor{gray}{TIP'24}} & 77.2 & 64.3 & 44.0 & 24.5 & 53.1 \\
AAPL \textsubscript{\textcolor{gray}{AAAI'25}} & 75.5 & 67.6 & 48.5 & 26.3 & 55.2 \\
TRA \textsubscript{\textcolor{gray}{ESWA'26}} & 83.3 & 74.5 & 58.9 & \underline{31.4} & 63.7 \\
PseR \textsubscript{\textcolor{gray}{TMM'26}} & \underline{88.9} & \underline{83.9} & \underline{66.1} & 28.9 & \underline{68.9} \\
Ours & \textbf{89.8} & \textbf{85.7} & \textbf{73.3} & \textbf{42.3} & \textbf{74.2} \\ \bottomrule
\end{tabular}

}
\end{table}

\section{Generalization across Proposal Generators and Detectors}

\label{sec:generalization}

\begin{figure}[t]
\centering
\includegraphics[width=\columnwidth]{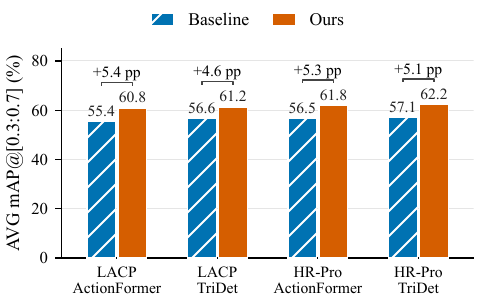}
\caption{\textbf{Generalization across proposal generators and detectors on THUMOS14.} For each pair, the baseline trains the detector on the generator's original pseudo-labels, while Ours applies Prior-Guided Initialization (PGI) and Consensus-Guided Refinement (CGR). AVG mAP is evaluated over tIoU thresholds 0.3--0.7. Brackets show gains in percentage points.}
\label{fig:generalization}
\end{figure}

On THUMOS14, we pair each proposal generator (LACP and HR-Pro) with each detector (ActionFormer and TriDet). As shown in Fig.~\ref{fig:generalization}, with LACP proposals, our framework improves AVG mAP over tIoU thresholds 0.3--0.7 by 5.4 and 4.6 percentage points for ActionFormer and TriDet, respectively. With HR-Pro proposals, the corresponding gains are 5.3 and 5.1 percentage points. The gains across all four pairs show that the framework works with both proposal generators and both detectors.

\section{Additional Ablation Studies}
\label{supp:ablations}
\subsection{Consistency-Controlled Refinement}

We isolate the role of consistency control in CGR on THUMOS14 by varying $\tau_{\mathrm{con}}^k$ while keeping PGI initialization and all other refinement settings fixed. This threshold determines whether a detector prediction agrees sufficiently with the current pseudo-label for an update. We compare no consistency constraint ($\tau_{\mathrm{con}}^k=0$), a fixed threshold ($\tau_{\mathrm{con}}^k=0.5$), and the progressive schedule that decreases the threshold from 1 to 0. The unconstrained variant updates every point with an available detector prediction. Detection performance is measured at the final training epoch, and pseudo-label quality is evaluated against official ground truth after the final update.

\begin{table}[t]
\centering
\caption{Ablation of consistency control in CGR on THUMOS14. AVG mAP (\%) is computed over tIoU thresholds 0.1--0.7. Both quality metrics compare final pseudo-labels with official ground truth over all training points, assigning zero tIoU to points without a pseudo-label.}
\label{tab:ablation_consistency}
\small
\setlength{\tabcolsep}{3pt}
\resizebox{\columnwidth}{!}{%
\begin{tabular}{lcccc}
\toprule
\multicolumn{1}{c}{\textbf{Consistency}} & \multirow{2}{*}{\textbf{\boldmath$\tau_{\mathrm{con}}^k$}} & \textbf{AVG mAP} & \textbf{Mean} & \textbf{\boldmath tIoU $\geq 0.7$} \\
\multicolumn{1}{c}{\textbf{control}} & & \textbf{(\%)} & \textbf{tIoU} & \textbf{(\%)} \\
\midrule
None & $0$ & 64.3 & 0.672 & 56.2 \\
Fixed & $0.5$ & 67.4 & 0.682 & 58.2 \\
\rowcolor{orange!12}
Progressive & $1\!\rightarrow\!0$ & 68.5 & 0.706 & 62.2 \\
\bottomrule
\end{tabular}

}
\end{table}

Table~\ref{tab:ablation_consistency} shows that progressive consistency control achieves 68.5\% AVG mAP, exceeding unconstrained updates and the fixed threshold by 4.2 and 1.1 percentage points, respectively. It also yields the highest mean pseudo-label tIoU (0.706) and fraction of points with tIoU $\geq 0.7$ (62.2\%). The latter exceeds the fixed-threshold result by 4.0 percentage points. These improvements support requiring strict agreement early in refinement and progressively allowing larger revisions as training proceeds.

\begin{figure}[t]
\centering
\includegraphics[width=\columnwidth]{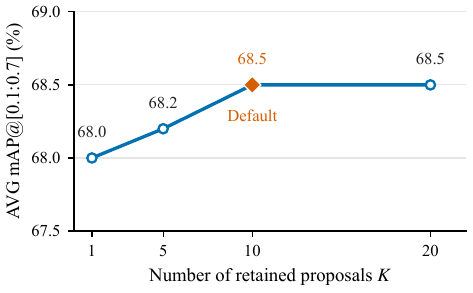}
\caption{Sensitivity to the number of retained proposals $K$ on THUMOS14, measured by AVG mAP (\%) over tIoU thresholds 0.1--0.7. The $K=1$ result corresponds to initial-boundary guidance. The orange diamond marks the default $K=10$.}
\label{fig:k_sensitivity}
\end{figure}

\begin{figure*}[t]
\centering
\includegraphics[width=\textwidth]{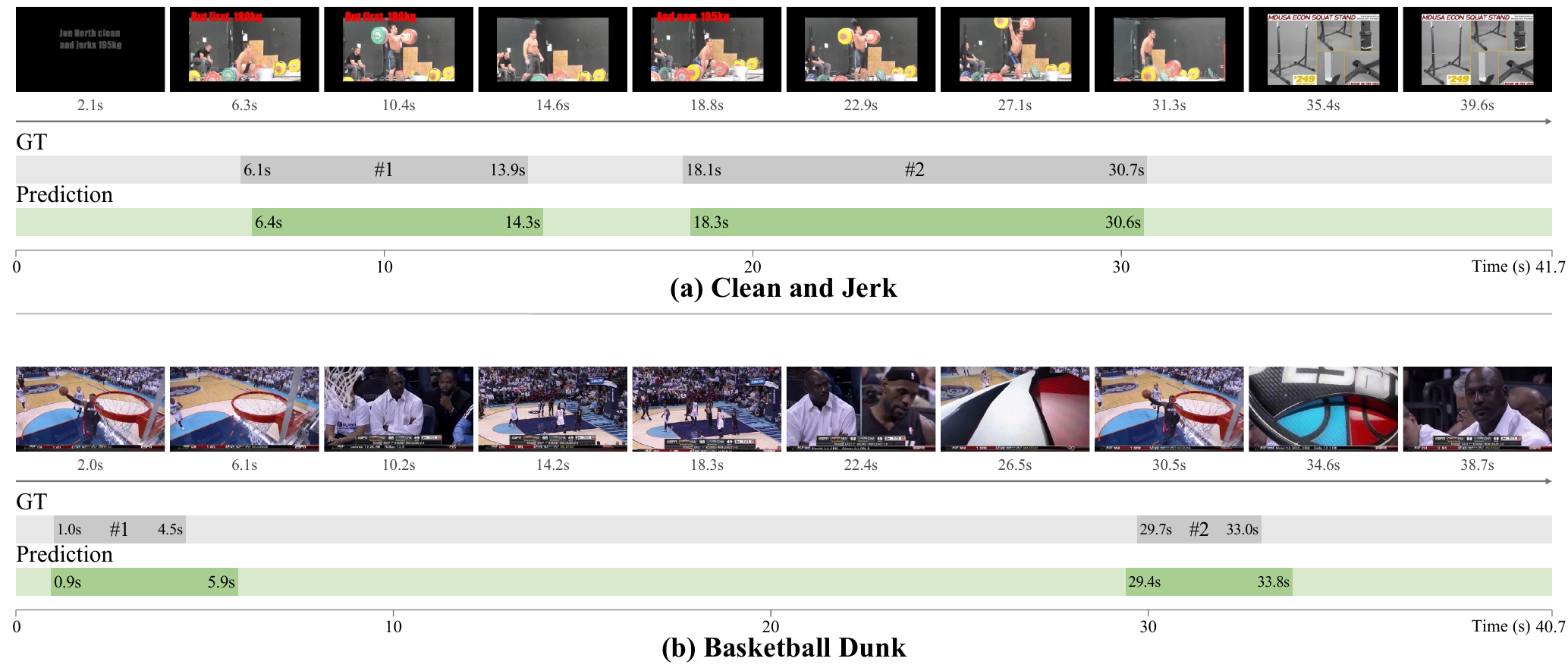}
\caption{\textbf{Final detection results on THUMOS14 test videos.} Ground-truth and predicted intervals are shown for Clean and Jerk and Basketball Dunk. All saved predictions with confidence $\geq 0.5$ are displayed.}
\label{fig:detection_qualitative}
\end{figure*}

\subsection{Number of Retained Proposals}

We vary the number of retained proposals $K\in\{1,5,10,20\}$ on THUMOS14 while keeping the initial PGI pseudo-labels and all other refinement settings fixed. With $K=1$, only the initial PGI boundary provides proposal evidence, corresponding to the initial-boundary variant in the main paper. Figure~\ref{fig:k_sensitivity} shows that AVG mAP increases from 68.0\% at $K=1$ to 68.2\% at $K=5$ and 68.5\% at $K=10$. Increasing $K$ to 20 yields the same AVG mAP, supporting $K=10$ as the default that retains fewer candidates without reducing detection performance.

\subsection{Consensus Fusion Threshold}
We vary $\tau_{\mathrm{iou}}\in\{0.1,0.3,0.5,0.7\}$ on THUMOS14 and ActivityNet1.3, keeping PGI initialization and all other refinement settings fixed within each dataset. This threshold controls whether detector--consensus agreement is sufficient to fuse the two intervals into a boundary target. Table~\ref{tab:boundary_sensitivity} shows that thresholds of 0.3 and 0.5 both achieve 68.5\% AVG mAP on THUMOS14 and 30.9\% on ActivityNet1.3. The default value of 0.5 also achieves 35.5\% mAP@0.7 on THUMOS14 and the highest mAP@0.95 of 7.2\% on ActivityNet1.3 among the tested settings. A threshold of 0.7 lowers both AVG mAP and mAP at the stricter evaluation thresholds on both datasets. We therefore use 0.5 as the shared default.

\begin{table}[t]
\centering
\caption{Sensitivity to the consensus fusion threshold $\tau_{\mathrm{iou}}$. All results are final-epoch mAP (\%). AVG denotes the average over tIoU thresholds 0.1--0.7 on THUMOS14 and 0.5--0.95 on ActivityNet1.3. The highlighted row uses the default threshold of 0.5.}
\label{tab:boundary_sensitivity}
\small
\setlength{\tabcolsep}{4pt}
\resizebox{\columnwidth}{!}{%
\begin{tabular}{ccccc}
\toprule
\multirow{2}{*}{\textbf{\boldmath$\tau_{\mathrm{iou}}$}} & \multicolumn{2}{c}{\textbf{THUMOS14}} & \multicolumn{2}{c}{\textbf{ActivityNet1.3}} \\
\cmidrule(lr){2-3}\cmidrule(lr){4-5}
 & \textbf{mAP@0.7} & \textbf{AVG} & \textbf{mAP@0.95} & \textbf{AVG} \\
\midrule
0.1 & 35.5 & 68.1 & 7.0 & 30.8 \\
0.3 & 35.1 & 68.5 & 7.1 & 30.9 \\
\rowcolor{orange!12}
0.5 & 35.5 & 68.5 & 7.2 & 30.9 \\
0.7 & 33.7 & 67.8 & 7.0 & 30.7 \\
\bottomrule
\end{tabular}

}
\end{table}

\section{Qualitative Detection Results}
\label{supp:detection_qualitative}

Figure~\ref{fig:detection_qualitative} compares final detector predictions with GT on two THUMOS14 test videos. For Clean and Jerk, the predicted start and end times closely match the annotated boundaries. For Basketball Dunk, the detector recovers two action instances separated by a long background interval. These examples illustrate the detector's ability to distinguish individual instances and localize their temporal extent.

\end{document}